\documentclass[runningheads]{llncs}
\usepackage{graphicx}
\usepackage{comment}
\usepackage{epsfig}
\usepackage{booktabs}
\usepackage{amsmath,amssymb} 
\usepackage{color}

\usepackage{amsfonts}       
\usepackage{nicefrac}       
\usepackage{color}
\usepackage{multirow}
\usepackage[super]{nth}
\usepackage{subfig}
\usepackage[normalem]{ulem} 
\usepackage{multirow, makecell}
\usepackage{xspace}
\usepackage{wrapfig}
\usepackage[pagebackref=true,breaklinks=true,letterpaper=true,colorlinks,bookmarks=false]{hyperref}

\newcommand{\myparagraph}[1]{\vspace{5pt}\noindent{\bf #1}}

\newcommand{\TASKFillIn}{Fill-in the Identity\xspace}
\newcommand{\TASKGenerate}{Identity-Aware Video Description\xspace}

\begin{document}
\pagestyle{headings}
\mainmatter
\def\ECCVSubNumber{3739}  

\title{Identity-Aware Multi-Sentence \\Video Description} 

\titlerunning{Identity-Aware Multi-Sentence \\Video Description}
\author{Jae Sung Park$^{1}$, Trevor Darrell$^{2}$, Anna Rohrbach$^{2}$}

\authorrunning{Park et al.}

\institute{Paul G. Allen School of Computer Science \& Engineering, University of Washington \and
University of California, Berkeley}
\maketitle

\setcounter{footnote}{0}

\begin{abstract}
Standard video and movie description tasks abstract away from person identities, thus failing to link identities across sentences. We propose a multi-sentence \TASKGenerate{} task, which overcomes this limitation and requires to re-identify persons locally within a set of consecutive clips. We introduce an auxiliary task of \TASKFillIn{}, that aims to predict persons' IDs consistently within a set of clips, when the video descriptions are given. Our proposed approach to this task leverages a Transformer architecture allowing for coherent joint prediction of multiple IDs. One of the key components is a gender-aware textual representation as well an additional gender prediction objective in the main model. This auxiliary task allows us to propose a two-stage approach to \TASKGenerate{}. We first generate multi-sentence video descriptions, and then apply our \TASKFillIn{} model to establish links between the predicted person entities. To be able to tackle both tasks, we augment the Large Scale Movie Description Challenge (LSMDC) benchmark with new annotations suited for our problem statement. Experiments show that our proposed \TASKFillIn{} model is superior to several baselines and recent works, and allows us to generate descriptions with locally re-identified people.\footnote{Project link at \href{https://sites.google.com/site/describingmovies/lsmdc-2019}{\color{blue}https://sites.google.com/site/describingmovies/lsmdc-2019}. \\ Code will be available at \href{https://github.com/jamespark3922/lsmdc-fillin}{\color{blue}{https://github.com/jamespark3922/lsmdc-fillin}}.}
\end{abstract}

\section{\label{sec:intro} Introduction}

\begin{figure}[t]
\scriptsize
\begin{center}
\includegraphics[width=\linewidth]{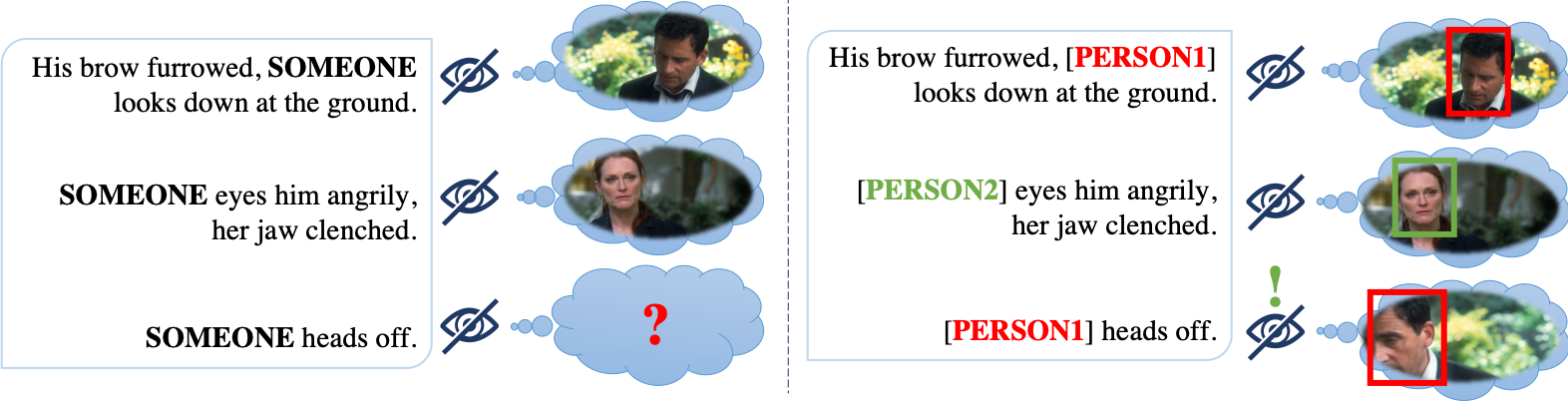}
\caption{Compare video description with SOMEONE labels vs. \textbf{\TASKGenerate{}}: in the first case it may be difficult for a visually impaired person to follow what is going on in the video, while in the second case it becomes clear who is performing which action.}
\vspace{-0.5cm}
\label{fig:teaser}
\end{center}
\end{figure}

Understanding and describing videos that contain multiple events often requires establishing ``who is who'': who are the participants in these events and who is doing what. Most of the prior work on automatic video description focuses on individual short clips and ignores the aspect of participants' identity. In particular, prior works on movie description tend to replace all character identities with a generic label SOMEONE \cite{torabi2015using,rohrbach16ijcv}. While reasonable for individual short clips, it becomes an issue for longer video sequences. As shown in Figure~\ref{fig:teaser}, descriptions that contain SOMEONE would not be satisfying to visually impaired users, as they do not unambiguously convey who is engaged in which action in video. 

Several prior works attempt to perform person re-identification~\cite{Taigman2014CvprDeepFace,Schroff2015ArxivFaceNet,cao2018vggface2} in movies and TV shows, sometimes relying on associated subtitles or textual descriptions \cite{tapaswi12cvpr,ramanathan14eccv,parkhi15}. Most such works take the ``linking tracks to names'' problem statement, i.e. trying to name all the detected tracks with proper character names. Others like \cite{pini2019mvad} aim to ``fill in'' the character proper names in the given ground-truth movie descriptions. 

\begin{figure}[t]
\scriptsize
\begin{center}
\includegraphics[width=0.75\linewidth]{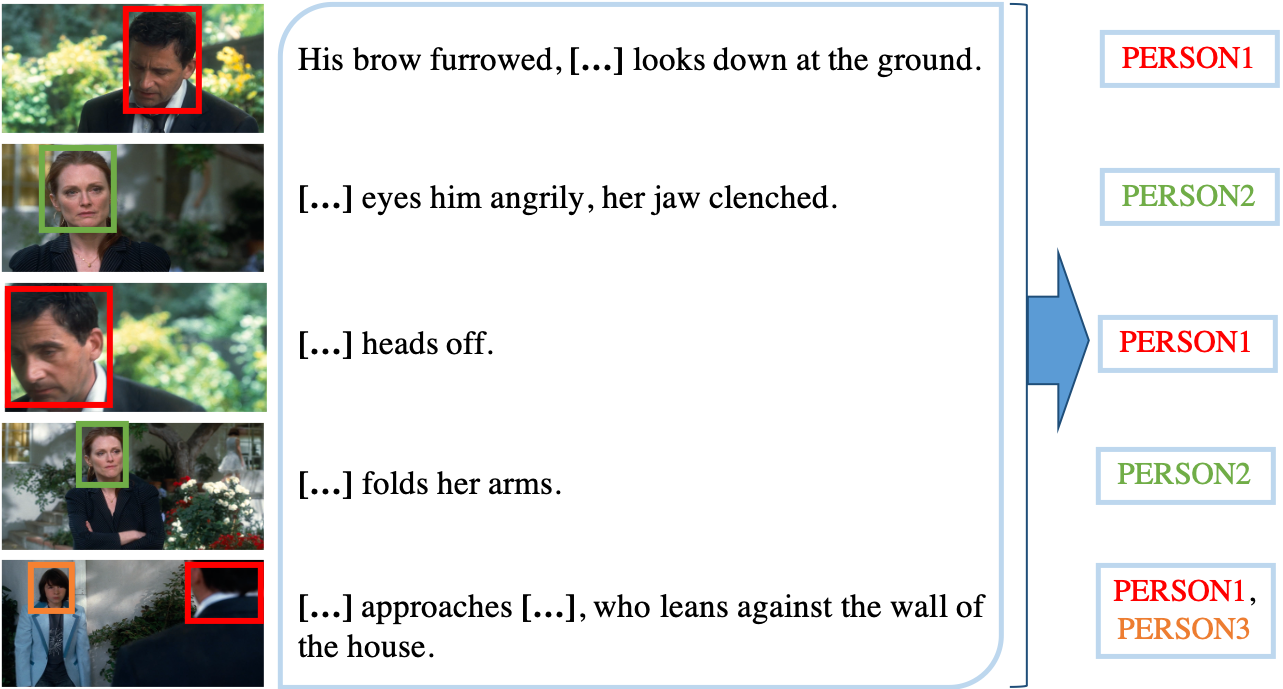}
\caption{Example of the \textbf{\TASKFillIn{}} task.}
\vspace{-0.8cm}
\label{fig:task2}
\end{center}
\end{figure}

In this work, we propose a different problem statement, which does not require prior knowledge of movie characters and their appearance. Specifically, we group several consecutive movie clips into sets and aim to establish person identities \emph{locally} within each set of clips. We then propose the following two tasks. First, given ground-truth descriptions of a set of clips, the goal is to fill in person identities in a coherent manner, i.e. to predict the same ID for the same person within a set (see Figure~\ref{fig:task2}). Second, given a set of clips, the goal is to generate video descriptions that contain corresponding local person IDs (see Figure~\ref{fig:teaser}). We refer to these two tasks as \textbf{\TASKFillIn{}} and \textbf{\TASKGenerate{}}. The first (auxiliary) task is by itself of interest, as it requires to establish person identities in a multi-modal context of video and description. 

We experiment with the Large Scale Movie Description Challenge (LSMDC) dataset and associated annotations~\cite{rohrbach16ijcv,rohrbach17cvpr}, as well as collect more annotations to support our new problem statement. We transform the global character information into local IDs within each set of clips, which we use for both tasks.

\myparagraph{\TASKFillIn{}.} Given textual descriptions of a sequence of events, we aim to fill in the person IDs in the blanks. In order to do that, two steps are necessary. First, we need to attend to a specific person in the video by relating visual observations to the textual descriptions. Second, we need to establish links within a set of blanks by relating corresponding visual appearances and textual context. We learn to perform both steps jointly, as the only supervision available to us is that of the identities, not the corresponding visual tracks. Our key idea is to consider an entire set of blanks jointly, and exploit the mutual relations between the attended characters. We thus propose a Transformer model~\cite{vaswani17nips} which jointly infers the identity labels for all blanks. Moreover, to support this process we make use of one additional cue available to us: the gender of the person in question. We train a text-based gender classifier, which we integrate in our model, along with an additional visual gender prediction objective, which aims to recognize gender based on the attended visual appearance. %

\myparagraph{\TASKGenerate{}.}
Given a set of clips, we aim to generate descriptions with local IDs. Here, we take a two-stage approach, where we first obtain the descriptions with SOMEONE labels as a first step, and next apply our \TASKFillIn{} method to give SOMEONEs their IDs. We believe there is a potential in exploring other models that would incorporate the knowledge of identities into generation process, and leave this to future work. 

Our contributions are as follows. (1) We introduce a new task of \textbf{\TASKGenerate{}}, which extends prior work in that it aims to obtain multi-sentence video descriptions with local person IDs. (2) We also introduce a task of \textbf{\TASKFillIn{}}, which, we hope, will inform future directions of combining identity information with video description. (3) We propose a Transformer model for this task, which learns to attend to a person and use gender evidence along with other visual and textual cues to correctly fill in the person's ID. (4) We obtain state-of-art results in the \textbf{\TASKFillIn{}} task, compared to several baselines and two recent methods. (5) We further leverage this model to address \textbf{\TASKGenerate{}} via a two-stage pipeline, and show that it is robust enough to perform well on the generated descriptions.
\section{\label{sec:related} Related Work}

\textbf{Video description.}
Automatic video description has attracted a lot of interest in the last few years, especially since the arrival of deep learning techniques \cite{donahue15cvpr,venugopalan15iccv,pan16cvpr,rohrbach15gcpr,yao2015iccv,youngjaeyu17cvpr,baraldi2017hierarchical,pan2017video}. Here are a few trends present in recent works ~\cite{wang2018bidirectional,li2018jointly,zhou2018end,rahman2019watch}. Some works formulate video description as a reinforcement learning problem~\cite{pasunuru2017reinforced,wang2018video,li2018end}. Several methods address grounding semantic concepts during description generation~\cite{zanfir16accv,wu2018interpretable,zhou2019grounded}. A few recent works put an emphasis on the use of syntactic information~\cite{wang2019controllable,hou2019joint}. New datasets have also been proposed~\cite{gella2018dataset,wang2019vatex}, including a work that focuses on dense video captioning~\cite{krishna2017dense}, where the goal is to temporally localize and caption individual events. %

In this work, we generate multi-sentence descriptions for long video sequences, as in \cite{rohrbach14gcpr,shin16icip,yu16cvpr}. This is different from dense video captioning, where one does not need to obtain one coherent multi-sentence description. Recent works that tackle multi-sentence video description include \cite{yu2018fine}, who generate fine-grained sport narratives, \cite{xiong2018move}, who jointly localize events and decide when to generate the following sentence, and \cite{park2019adversarial}, who introduce a new inference method that relies on multiple discriminators to measure the quality of multi-sentence descriptions. %

\textbf{Person re-identification.}
Person re-identification aims to recognize whether two images of a person are the same individual. This is a long standing problem in computer vision, with numerous deep learning based approaches introduced over the years~\cite{Taigman2014CvprDeepFace,parkhi15bmvc,Schroff2015ArxivFaceNet,wang2017normface,cao2018vggface2}. We rely on \cite{Schroff2015ArxivFaceNet} as our face track representation.

\textbf{Connections to prior work.}
Next, we detail how our work compares to the most related prior works.

\emph{\TASKGenerate{}:} Closely related to ours is the work of \cite{rohrbach17cvpr}. They address video description of \emph{individual} movie clips with grounded and co-referenced (re-identified) people. In their problem statement re-identification is performed w.r.t. \emph{a single previous clip}  during description generation. Unlike \cite{rohrbach17cvpr}, we address \emph{multi-sentence} video description, which requires consistently re-identifying people over \emph{multiple clips at once} (on average 5 clips).

\emph{\TASKFillIn{}:} Our task of predicting \emph{local} character IDs for a set of clips given ground-truth descriptions with blanks, is related to the work of \cite{pini2019mvad}. However, they aim to fill in \emph{global} IDs (proper names). In order to learn the global IDs, they use {80\%} of each movie for training. Our problem statement is different, as it requires no access to the movie characters' appearance during training: we maintain disjoint training, validation and test movies. A number of prior works attempt to link all the detected face tracks to global character IDs in TV shows and movies~\cite{Everingham06bmvc,sivic09cvpr,tapaswi12cvpr,bojanowski13iccv,ramanathan14eccv,parkhi15,miech2017learning,jin2017end}, which is different from our problem statement that tries to fill character IDs locally with textual guidance. We compare to two recent approaches to \TASKFillIn{} in Section~\ref{sec:exp:task2:test}.

\section{\label{sec:approach} Connecting Identities to Video Descriptions}

\begin{figure*}[t]
\scriptsize
\begin{center}
\includegraphics[width=\linewidth]{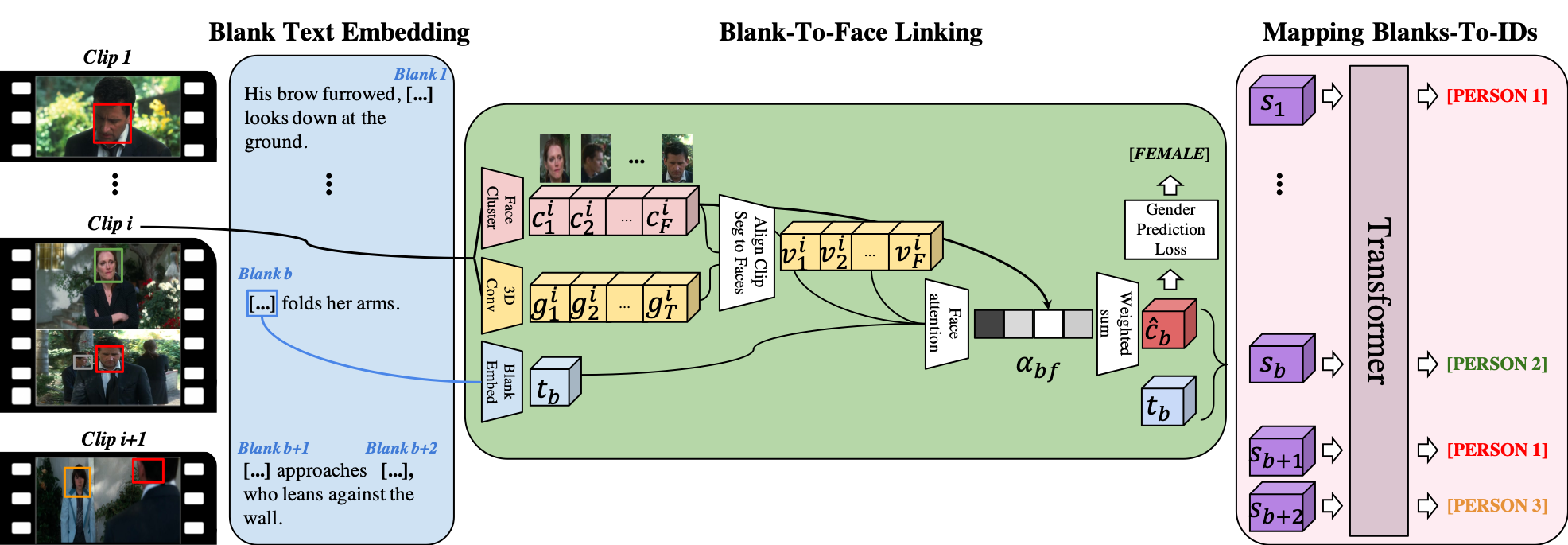}
\caption{Overview of our approach to \textbf{\TASKFillIn{}} task. See Section~\ref{ssec:fillin}.}
\vspace{-0.5cm}
\label{fig:approach:task2}
\end{center}
\end{figure*}

An integral part of understanding a story depicted in a video is to establish who are the key participants and what actions they perform over the course of time. Being able to correctly link the repeating appearances of a certain person could potentially help follow the story line of this person. We first address the task of \textbf{\TASKFillIn{}}, where we aim to solve a related problem: fill in the persons' IDs based on the given video descriptions with blanks. Our approach is centered around two key ideas: joint prediction of IDs via a Transformer architecture~\cite{vaswani17nips}, supported by gender information inferred from textual and visual cues (see Figure~\ref{fig:approach:task2}). We then present our second task, \textbf{\TASKGenerate{}}, which aims to generate multi-sentence video descriptions with local person IDs. We present a two-stage pipeline, where our baseline model gives us multi-sentence descriptions with SOMEONE labels, after which we leverage the \textbf{\TASKFillIn{}} auxiliary task to link the predicted SOMEONE entities. 

\subsection{\label{ssec:fillin} \TASKFillIn{}}

For a set of video clips $V_i$ and their descriptions with blanks $D_i, i={1,2...,N}$, we aim to fill in character identities that are locally consistent within the set. We first detect faces that appear in each clip and cluster them based on their visual appearance. Then, for every blank in a sentence, we attend over the face cluster centers using visual and textual context, to find the cluster best associated with the blank. We process all the blanks sequentially and pass their visual and textual representations to a Transformer model \cite{vaswani17nips}, which analyzes the entire set at once and predicts the most probable sequence of character IDs (Figure~\ref{fig:approach:task2}). 

\paragraph{Visual Representation.}
Now, we describe the details of getting local face descriptors and other global visual features for a clip $V_i$. We detect all the faces in every frame of the clip, using the face detector by \cite{zhang2016mtcnn}. Then, we extract 512-dim face feature vectors with the FaceNet model \cite{Schroff2015ArxivFaceNet}\footnote{https://github.com/davidsandberg/facenet} trained on the VGGFace2 dataset \cite{cao2018vggface2}. The feature vectors are clustered using DBSCAN algorithm \cite{ester1996dbscan}, which does not require specifying a number of clusters as a parameter. We take the mean of face features in each cluster, resulting in $F$ face feature vectors $(c^i_1, ..., c^i_F)$.

In addition to the face features, we extract spatio-temporal features that describe the clip semantically. These features help the model locate where to look for the relevant face cluster for a given blank. We extract I3D~\cite{carreira2017quo} %
features and apply mean pooling across a fixed number of $T$ segments following \cite{wang2016temporal}%
, giving us a sequence $(g^i_1, ..., g^i_T)$. We then associate each face cluster with the best temporally aligned segment as follows. For each face cluster $c^i_f$, we keep track of its frame indices and get a ``center'' index. Dividing this index by the total number of frames in a clip gives us a relative temporal position of the cluster $r^i_f$, $0 <= r^i_f < 1$. We get the corresponding segment index $t^i_f = \lfloor{r^i_f * T}\rfloor$ and obtain the global visual context $v^i_f = g^i_{t^i_f}$. We concatenate face cluster features $c^i_f$ with the associated global visual context $v^i_f$ as our final visual representation.

\emph{Filling in the Blanks.}
Suppose there are $B$ blanks in the set of $N$ sentences ${D_i}$\footnote{Some sentences may have multiple blanks, others may have none.}. One way to fill in these blanks is to train a language model, such as BERT~\cite{devlin2018bert}, by masking the blanks to directly predict the character IDs. As we aim to incorporate visual information in our model, we take the following approach. 

First, each blank $b$ in a sentence $D_i$ has to receive a designated textual encoding.  %
We use with %
a pretrained BERT model, which has been shown effective for numerous NLP tasks. Instead of relying on a generic pretrained model, we train it to predict the \emph{gender} corresponding to each blank, which often can be inferred from text. For example, in Figure~\ref{fig:task2}, one can infer that the person in the first clip is male due to the phrase ``His brow''. %
We process all sentences in the set jointly. To get a representation for each blank, we access output embedding from the [CLS] token, a special sentence classification token in \cite{devlin2018bert} whose representation captures the meaning of the entire sentence, over all sentences, and a hidden state of the last layer associated with the specific blank token. Note, that the same [CLS] token is used for all blanks in the set. The final representation $t_b$ is a concatenation of the [CLS] and the blank token representation.

For each clip $V_i$, we obtain $F$ face cluster representations $(c^i_1, ... c^i_F)$, which we combine with the corresponding clip level representations $(v^{i}_{1}, ... v^{i}_{F})$. To find the best matching face cluster for the blank $b$, we predict the attention weights $\alpha_{bf}$ over all clusters in the clip based on the $t_b$, and compute a weighted sum over the face clusters, $\hat{c_b}$: %
\begin{equation}
\begin{aligned} 
e_{bf} = W_{\alpha_2}\tanh(W_{\alpha_1}[c^i_f;v^i_f;t_b]), \\
\alpha_{bf} = \frac{\exp(e_{bf})}{\sum_{k=1}^F{\exp(e_{bk})}}, \hat{c_b} = \sum_{f=1}^F{\alpha_{bf}c_f}
\end{aligned}
\label{eq:face_attention}
\end{equation}
We concatenate the visual representation $\hat{c_b}$ with $t_b$ as the final representation for the blank: $s_b = [\hat{c_b}; t_b]$. 

Given a set of $B$ blanks represented by $(s_1, ..., s_B)$ we now aim to link the corresponding identities to each other. Instead of making pairwise decisions w.r.t. matching and non-matching blanks, we want to predict an entire sequence of IDs jointly. We thus choose the Transformer \cite{vaswani17nips} architecture to let the self-attention mechanisms model multiple pairwise relationships at once. Specifically, we train a Transformer with $(s_1, ..., s_B)$ as inputs and local person IDs $(l_1, ..., l_B)$ as outputs. As we fill in the blanks in a sequential manner, we prevent the future blanks from impacting the previous blanks by introducing a causal masking in the encoder. 
We train the entire model end-to-end and learn the attention mechanism in Eq.~\ref{eq:face_attention} jointly with the ID prediction. Denoting $\theta$ as all the parameters in the model, the loss function we minimize is:
\begin{equation}
\begin{aligned} 
L_{character}(\theta) = - \sum_{b=1}^B {p_\theta(l_b \mid s_1, ..., s_{b-1}, l_1, ..., l_{b-1})} 
\end{aligned}
\end{equation}

We explore the effect of an additional component, a \emph{gender prediction loss} $L_{gender}$, that forces the attended visual representation to be gender-predictive. We add a single layer perceptron that takes the predicted feature $\hat{c_b}$ and aims to recognize the gender $g_b$ for the blank $b$. The final loss function we minimize is as follows:

\begin{equation}
\begin{aligned} 
L_{gender}(\theta) = - \sum_{b=1}^B {p_\theta(g_b \mid \hat{c_b})} \\ 
L(\theta) = L_{character} + \lambda_{gen} L_{gender}
\end{aligned}
\end{equation}
where $\hat{c_b}$ is calculated in Eq~\ref{eq:face_attention} and $\lambda_{gen}$ is a hyperparameter. 

We also notice that it is possible to boost the performance of our Transformer model by a simple training data augmentation. Note that there are various ways to split the training data into clip sequences with length $N$: one can consider a non-overlapping segmentation, i.e. $\{1,...N\}, \{N+1,...2N\}, ...$ or additionally add all the overlapping sets $\{2,...N+1\}, \{3,...N+2\}, ...$ . Since we predict the local IDs, every such set would result in a unique data point, meaning using all the possible sets can increase the amount of training data by a factor of $N$.

\subsection{\TASKGenerate{}\label{ssec:generate}}
Here, given a set of $N$ clips $V_i$, we aim to predict their descriptions $D_i$ that would also contain the relevant local person IDs. 
First, we follow prior works \cite{park2019adversarial,gella2018dataset} to build a multi-sentence description model with SOMEONE labels. It is an LSTM based decoder that takes as input a visual representation and a sentence generated for a previous clip. Here, our visual representation for $V_i$ is %
I3D \cite{carreira2017quo} and Resnet-152~\cite{he2016deep} mean pooled temporal segments. Once we have obtained a multi-sentence video description with SOMEONE labels, we process the generated sentences with our \TASKFillIn{} model. We demonstrate that this approach is effective, although the \TASKFillIn{} model is only trained on ground-truth descriptions. Note, that this two-stage pipeline could be applied to any video description method.

\section{\label{sec:dataset} Dataset}

As our main test-bed, we choose the Large Scale Movie Description Challenge (LSMDC)~\cite{rohrbach16ijcv}, while leveraging and extending the character annotations from~\cite{rohrbach17cvpr}. They have marked every sentence in the MPII Movie Description (MPII-MD) dataset where a person's proper name (e.g. \emph{Jane}) is mentioned, and labeled the person specific pronouns \emph{he, she} with the associated names (e.g. \emph{he} is \emph{John}). For each one out of 94 MPII-MD movies, we are given a list of all unique identities and their genders. We extend these annotations to 92 additional movies, covering the entire LSMDC dataset (except for the Blind Test set). 

\newcommand{\midruleStats}{\cmidrule(rr){1-1} \cmidrule(rr){2-2} \cmidrule(rr){3-5}}
\begin{table}[t]
\center
\footnotesize
\scalebox{0.9}{
\begin{tabular}{l@{\ }r@{\ \ \ }r@{\ \ \ }r@{\ \ \ }r}
\toprule
           & Movies & Sentences & Sets & Blanks  \\
\midruleStats
Training   & 153 & 101,079 & 20,283 & 87,604 \\
Validation & 12 &  7,408 &  1,486  &  6,457 \\
Public Test       & 17 & 10,053 &  2,018  &  8,431  \\
\bottomrule
\end{tabular}
}
\vspace{0.5cm}
\caption{Statistics for our tasks, based on the LSMDC dataset. See Section~\ref{sec:dataset}.}
\vspace{-0.5cm}
\label{tbl:task_stat}
\end{table}

We use these annotations as follows.
(1) We drop the pronouns and focus on the underlying IDs. (2) We split each movie into sets of consecutive 5 clips (the last set in a movie may  contain less than 5 clips). (3) We relabel global IDs into local IDs \emph{within each set}. E.g. if we encounter a sequence of IDs \emph{Jane, John, Jane, Bill}, it will become \emph{PERSON1, PERSON2, PERSON1, PERSON3}. This relabeling is applied for both tasks that we introduce in this work.

We provide dataset statistics, including number of movies, individual sentences, sets and blanks in Table~\ref{tbl:task_stat}\footnote{Note, that the reported number of training clip sets reflects the default non-overlapping ``segmentation'', as done for validation and test movies. One is free to define the training clip sets as arbitrary sets of 5 consecutive clips.}. Around {52\%} of all blanks correspond to PERSON1, {31\%} -- to PERSON2, {12\%} -- to PERSON3, {4\%} -- to PERSON4, {1\%} or less -- to PERSON5, PERSON6, etc. This shows that the clip sets tend to focus on a single person, but there is still a challenge in distinguishing PERSON1 from PERSON2, PERSON3, ... (up to PERSON11 in training movies). 

In our experiments we use the LSMDC Validation set for development, and LSMDC Public Test set for final evaluation.

\section{\label{sec:experiments} Experiments}

\subsection{Implementation details}

For each clip, we extract I3D~\cite{carreira2017quo} pre-trained on the Kinetics~\cite{kay2017kinetics} dataset and Resnet-152~\cite{he2016deep} features pre-trained on the ImageNet~\cite{deng2009imagenet} dataset. We mean pool them temporally to get $T=5$ segments \cite{wang2016temporal}. We detect on average 79 faces per clip (and up to 200). In the DBSCAN algorithm used to cluster faces, we set $\epsilon = 0.2$, which is the maximum distance between two samples in a cluster. Clustering the faces within each clip results in about 2.2 clusters per clip, and clustering over a set results in 4.2 clusters per set. BERT~\cite{devlin2018bert} models use the BERT-base model architecture with default settings as in \cite{Wolf2019HuggingFacesTS}. %
Transformer~\cite{vaswani17nips} has a feedforward dimension of 2048 and 6 self-attention layers. We train the \TASKFillIn{} model for 40 epochs with learning rate 5e-5 with hyperparameter $\lambda_{gender} = 0.2$. We train the baseline video description model for 50 epochs with learning rate 5e-4. We fix batch size as 16 across all experiments, where each batch contains a set of clips and descriptions. %

\subsection{\label{sec:exp:task2} \textbf{\TASKFillIn{}}}

\subsubsection{Evaluation metrics}
First, we discuss the metrics used to evaluate the \textbf{\TASKFillIn{}} task. Given a sequence of blanks and corresponding ground-truth IDs, we consider all unique pairwise comparisons between the IDs. A pair is labeled as ``Same ID'' if the two IDs are the same, and ``Different ID'' otherwise. We obtain such labeling for the ground-truth and predicted IDs. Then, we can compute a ratio of the matching labels between the ground-truth and predicted pairs, e.g. if 6 out of 10 pair labels match, the prediction gets an accuracy {0.6}\footnote{This resembled pairwise precision/recall used in clustering~\cite{banerjee2005model}. However, these are not applicable in our scenario as they can not handle singleton clusters (with one element). Thus, we compute pairwise accuracy instead.}. The final accuracy is averaged across all sets. When define like this, we obtain an \emph{instance-level} accuracy over ID pairs (``Inst-Acc''). Note, that it is important to correctly predict both ``Same ID'' and ``Different ID'' labels, which can be seen as a 2-class prediction problem. The instance-level accuracy does not distinguish between these two cases. Thus, we introduce a \emph{class-level} accuracy, where we separately compute accuracy over the two subsets of ID pairs (``Same-Acc'', ``Diff-Acc'') and report the harmonic mean between the two (``Class-Acc'').

\subsubsection{Baselines and ablations}
\label{sec:exp:task2:val}

Table~\ref{tab:task2_val_new} summarizes our experiments on the LSMDC Validation set. We include two simple baselines: ``The same ID'' (all IDs are the same) and ``All different IDs'' (all IDs are distinct: 1, 2, ...). ``GT Gender as ID'' directly uses \emph{ground truth} male/female gender as a character ID (Person 1/2), and serves as an upper-bound for gender prediction. We consider two vision-only baselines, where we cluster all the detected faces within a set of 5 clips, and pick a random cluster (``Random Face Cluster'') or the most frequent cluster (``Most Frequent Face Cluster'') within a clip for each blank. We also consider a language-only baseline ``BERT Character LM'', which uses a pretrained BERT model to directly fill in all the blanks.
Then we include our Transformer model with %
our BERT Gender pretrained model. We show the effect of our training data augmentation and use augmentation in the following versions. Finally, we study the impact of adding each visual component (``+ Face.'' and ``+ Video''), and introduce our vision-based gender loss (full model).

\newcommand{\midruleFIB}{\cmidrule(rr){1-1} \cmidrule(rr){2-4} \cmidrule(rr){5-5} \cmidrule(rr){6-6}}
\begin {table}[t]
\begin{center}
\footnotesize
\scalebox{0.8}{
\begin{tabular}{@{}l@{}c@{}@{}c@{}c@{}c@{}c@{}c@{}}
\toprule
\textbf{} & \textbf{  Same } & \textbf{  Diff } & \textbf{  Class } & \textbf{   Inst } &\textbf{  Gen}\\
\textbf{Method} & \textbf{  Acc } & \textbf{  Acc } & \textbf{  Acc } & \textbf{   Acc } &\textbf{  Acc}\\
\midruleFIB
The same ID & 100.0 & 0.0 & 0.0 & 40.7 &-  \\
All different IDs & 0.0 & 100.0 & 0.0 & 59.3 & - \\
\midruleFIB
GT Gender as ID & 100.0 & 43.0 & 60.1 & 65.5 & - \\
\midruleFIB
V: Random Face Cluster & 54.8 & 52.0 & 53.4&  55.3 & -\\
V: Most Frequent Face Cluster &  48.3&  68.8 & 56.7 & 61.6 & -\\
\midruleFIB
L: BERT Character LM & 57.9 & 65.1 & 57.9& 66.4 & -\\
\midruleFIB
L: Transf. + BERT Gen. LM & 57.3 & 68.9 & 62.6 & 67.9 & 80.3 \\ %
L: Transf. + BERT Gen. LM + Aug. & 60.7 & 68.7 & 64.4 & 69.6 & 81.8 \\ %
L+V: Transf. + BERT Gen. LM + Aug. + Face. &  62.8 & 68.0 & 65.3 & 69.7 & 81.8 \\ %
L+V: Transf. + BERT Gen. LM + Aug. + Face + Video. & 64.8 & 66.6 & 65.7 & 69.6 & 81.8 \\
L+V: Transf. + BERT Gen. LM + Aug. + Face + Video + Gen. Loss & 63.5 & 68.4 & 65.9 & 69.8 & 83.0 \\
\bottomrule
\end{tabular}}
\end{center}
\caption{\textbf{\TASKFillIn{}} accuracy of several baselines, our full method and its ablations on the LSMDC Validation set. We report the predicted ID accuracy at class and instance level, as well as gender accuracy. See Section~\ref{sec:exp:task2:val} for details.} 
\vspace{-1cm}
\label{tab:task2_val_new}
\end {table}

We make the following observations. 
(1) Instance accuracy for all the same/all distinct IDs provides the insight into how the pairs are distributed ($40.7\%$ of all pairs belong to ``Same ID'' class, $59.3.7\%$ -- to ``Different ID'' class). Neither is a good solution, getting Class-Acc $0$.
(2) Our Transformer model with BERT Gender representation improves over the vanilla BERT Character model ({57.9} vs. {62.6} in Class-Acc). (3) This is also higher than {60.1} of ``GT Gender as ID'', i.e. our model relies on other language cues besides gender. (4) Training with our data augmentation scheme further improves Class-Acc to {64.4}. (5) Introducing face features boosts the Class-Acc to {65.3}, and adding video features improves it to {65.7}. (7) Finally, visual gender prediction loss %
leads to the overall Class-Acc of {65.9}. Note, that the instance-level accuracy (Inst-Acc) does not always reflect the improvements as it may favor the majority class (``Different ID'').

We also report gender accuracy (Gen Acc) for the variants of our model (last 4 rows in Table~\ref{tab:task2_val_new}). For models without the visual gender loss, we report the accuracy based on their BERT language model trained for gender classification. %
We see that data augmentation on the language side helps improve gender accuracy ({80.3} vs {81.8}). Incorporating visual representation with the gender loss boosts the accuracy further ({81.8 vs 83.0}).

\newcommand{\midruleFIBTest}{\cmidrule(rr){1-1} \cmidrule(rr){2-4} \cmidrule(rr){5-5}}
\begin{wraptable}{r}{0.5\linewidth}
\renewcommand{\arraystretch}{0.98}
\footnotesize
\vspace{-1.2cm}
\begin{centering}
\begin{tabular}{@{}l@{}c@{}@{}c@{}c@{}c@{}c@{}}
\toprule
\textbf{} & \textbf{  Same } & \textbf{  Diff } & \textbf{  Class } & \textbf{  Inst }\\
\textbf{Method} & \textbf{  Acc } & \textbf{  Acc } & \textbf{  Acc } & \textbf{  Acc }\\
\midruleFIBTest
Human w/o video & 56.5 & 78.5 & 65.7 & 70.0 \\ 
Human & 83.9 & 90.0 & 86.8 & 87.0 \\
\midruleFIBTest
Ours & 64.6 & 70.7 & 67.5 & 70.3 \\
\bottomrule
\end{tabular}
\par\end{centering}
\caption{\textbf{\TASKFillIn{}} human performance and our method evaluated on 200 random Test clips sets.} 
\vspace{-0.5cm}
\label{tab:task2_test_human}
\end{wraptable}

\subsubsection{Human performance}\label{sec:exp:task2:human}
We also assess human performance in this task in two scenarios: with and without seeing the video. The former provides an overall upper-bound accuracy, while the latter gives an upper-bound for the text-only models. We randomly select 200 sets of Test clips and ask 3 different Amazon Mechanical Turk (AMT) workers to assign the IDs to the blanks. For each set we compute a \emph{median} accuracy across the 3 workers to remove the outliers and report the average accuracy over all sets. Table~\ref{tab:task2_test_human} reports the obtained human performance and the corresponding accuracy of our model on the same 200 sets. %
Human performance gets a significant boost when the workers can see the video, indicating that \emph{video provides many valuable cues}, and not everything can be inferred from text. Our full method outperforms ``Human w/o video'' but falls behind the final ``Human'' performance.
\vspace{-0.5cm}
\subsubsection{Comparison to state-of-the-art}\label{sec:exp:task2:test}
Since the data for our new tasks has been made public, other research groups have reported results on it, including the works by Yu et al.~\cite{yu19lsmdc} and Brown et al~\cite{brown19lsmdc}\footnote{Note, that we have corrected some errors, affecting about $3\%$ of the annotations. While Yu et al.~\cite{yu19lsmdc} and Brown et al~\cite{brown19lsmdc} have trained their models on the old annotations, all reported results are obtained on the \emph{corrected} test set.}.
Yu et al. propose an ensemble over two models: Text-Only and Text-Video. The Text-Only model %
builds a pairwise matrix for a set of blanks and learns to score each pair by optimizing a binary cross entropy loss. 
The Text-Video model considers two tasks: linking blanks to tracks and linking tracks to tracks. The two tasks are trained separately, with the help of additional supervision from an external dataset~\cite{pini2019mvad}, using triplet margin loss. %
While Yu et. al. use gender loss to pre-train their language model, we introduce gender loss on the visual side.

\begin{wraptable}{r}{0.5\linewidth}
\renewcommand{\arraystretch}{0.98}
\footnotesize
\vspace{-0.5cm}
\begin{centering}
\begin{tabular}{@{}l@{}c@{}@{}c@{}c@{}c@{}c@{}}
\toprule
\textbf{} & \textbf{  Same } & \textbf{  Diff } & \textbf{  Class } & \textbf{  Inst }\\
\textbf{Method} & \textbf{  Acc } & \textbf{  Acc } & \textbf{  Acc } & \textbf{  Acc }\\
\midruleFIBTest
Yu et al.~\cite{yu19lsmdc} & 26.4 & 87.3 & 40.6 & 65.9 \\
Brown et al.~\cite{brown19lsmdc} & 33.6 & 81.0 & 47.5 & 64.8 \\
\midruleFIBTest
Ours text only & 56.0 & 71.2 & 62.7 & 68.0 \\
Ours & 60.6 & 70.0 & 64.9 & 69.6  \\
\bottomrule
\end{tabular}
\par\end{centering}
\caption{\textbf{\TASKFillIn{}} accuracy of our method and two recent works on the LSMDC Test set.} 
\vspace{-0.5cm}
\label{tab:task2_test_sota}
\end{wraptable}

Brown et al. %
train a Siamese network on positive (same IDs) and negative (different IDs) pairs of blanks. The network relies on an attention mechanism over the face features to identify the relevant person given the blank encoding.

\begin{figure}[t]
\scriptsize
\begin{center}
\includegraphics[width=0.9\linewidth]{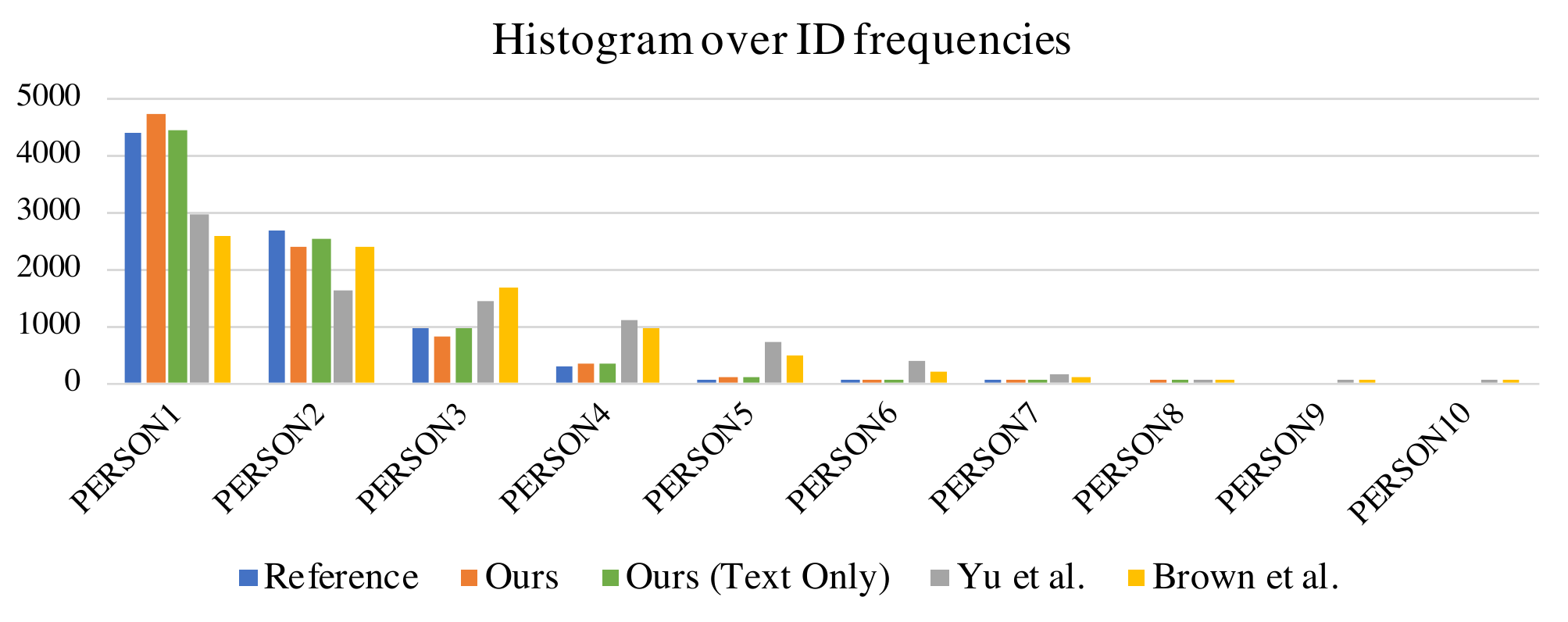}
\caption{\textbf{\TASKFillIn{}}: histogram over the frequencies of predicted IDs for our method, its text-only version and two SOTA works. See Section~\ref{sec:exp:task2}.}
\vspace{-0.7cm}
\label{fig:hist_task2}
\end{center}
\end{figure}

Table~\ref{tab:task2_test_sota} reports the results on the LSMDC Test set. As we see, our approach significantly outperforms both methods. To gain further insights, we analyze some differences in behavior between these methods and our approach.

\begin{figure}[t]
\scriptsize
\begin{center}
\includegraphics[width=\linewidth]{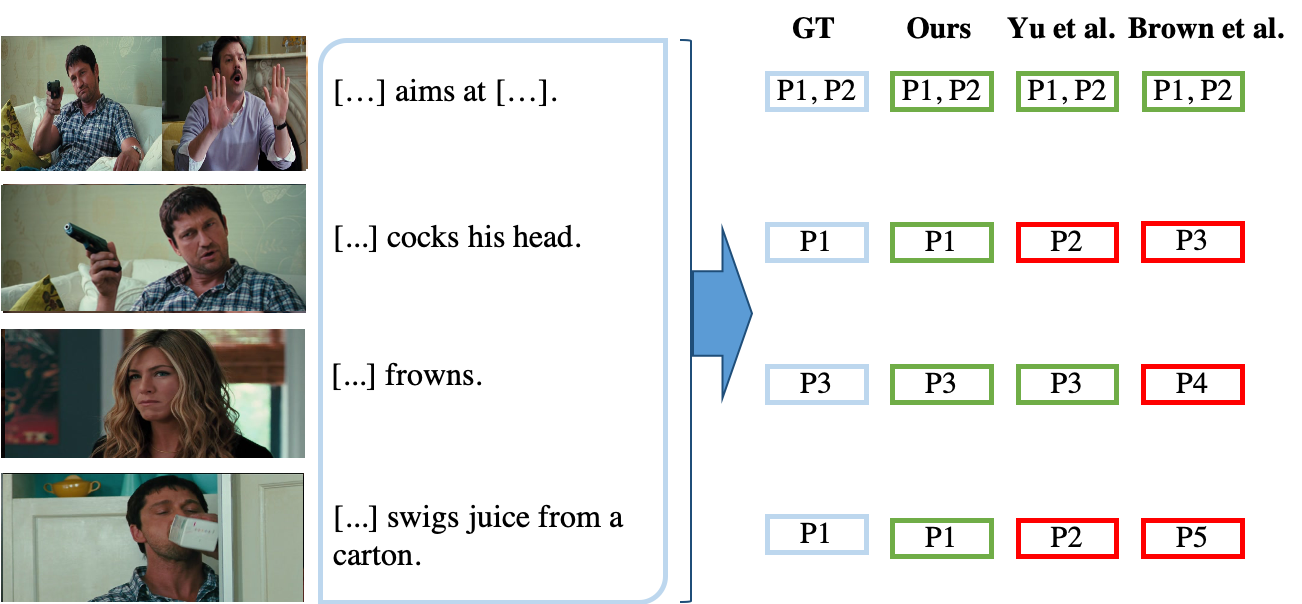}
\caption{Qualitative example for \textbf{\TASKFillIn{}} task, comparison between our approach and two recent methods. Correct/incorrect predictions are labeled with green/red, respectively. P1, P2, ... are person IDs. See Section~\ref{sec:exp:task2:qual} for details.}
\vspace{-0.5cm}
\label{fig:qual_task2_sota}
\end{center}
\end{figure}

\begin{figure}[t]
\scriptsize
\begin{center}
\includegraphics[width=0.9\linewidth]{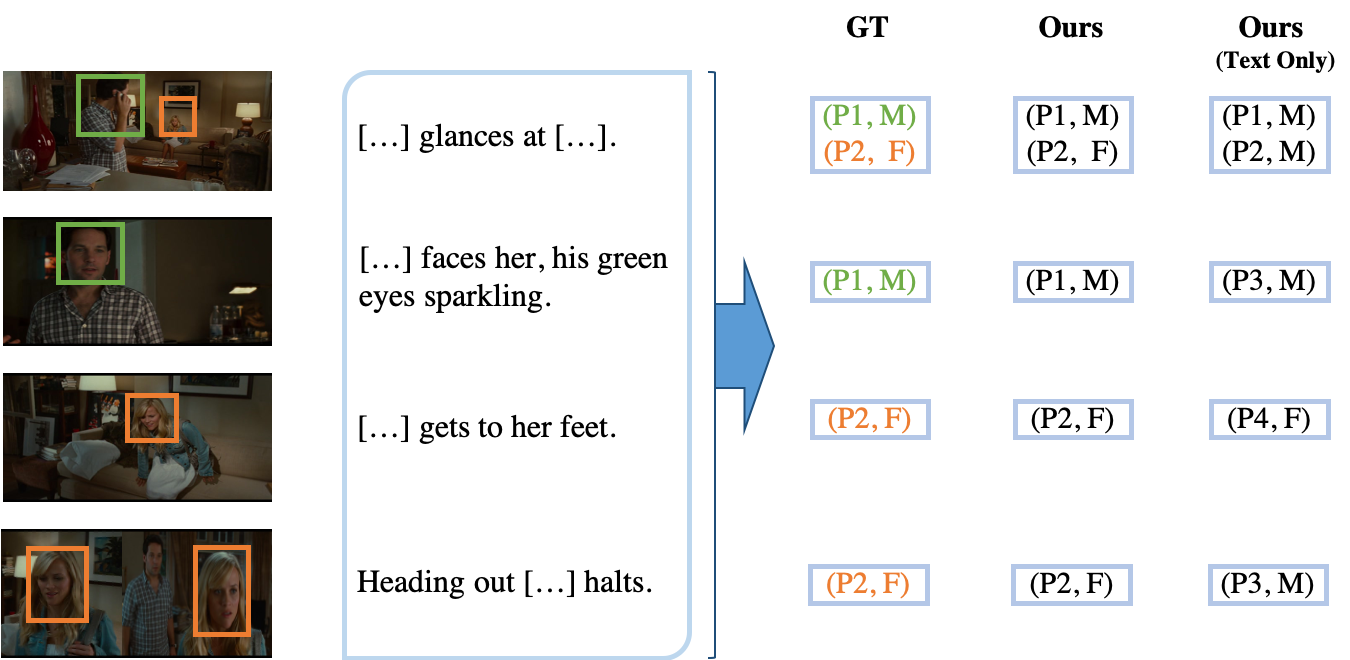}
\caption{Qualitative example for \textbf{\TASKFillIn{}} task, comparison between our final approach with visual representation and a text-only ablation (``Transf. + BERT Gender LM + Augm.'' in Table~\ref{tab:task2_val_new}). We include the predicted character ID (P1, P2, ...) and gender for each blank. See Section~\ref{sec:exp:task2:qual} for details.}
\vspace{-0.5cm}
\label{fig:qual_task2_ablation}
\end{center}
\end{figure}

We take a closer look at the distribution of the predicted IDs (PERSON1, PERSON2, ...) by our method, and the two other approaches. Figure~\ref{fig:hist_task2} provides a histogram over the reference data and the compared approaches. We can see that our predictions align well with the true data distribution, while the two other methods struggle to capture the data distribution. Notably, both methods favor more diverse IDs ({2, 3, ...}), failing to link many of the re-occurring character appearances. This may be in part due to the difference in the objective used in our approach vs. the others. While they implement a binary classifier that selects the best matching face track for each blank, we use Transformer to fill in the blanks jointly, allowing us to better capture both local and global context. 

Figure~\ref{fig:qual_task2_sota} provides a qualitative example, comparing our approach, Yu et al. and Brown et al. As suggested by our analysis, these two methods often predict diverse IDs instead of recognizing the true correspondences. %

\subsubsection{Ours vs. Ours Text Only}\label{sec:exp:task2:qual}

In Figure~\ref{fig:qual_task2_ablation}, we compare our full model vs. its text-only version. After having seen the two characters in the first clip (P1 and P2), our full model recognizes that the man and woman appearing in a next set of clips are the same two characters, and successfully links them as P1 in the second and P2 in the third and fourth clip with the correct genders. In the last clip with two characters, the full model is also able to visually ground the same woman as ``heading out'' and assign the ID as P2. On the other hand, the text-only model cannot tell that the first two characters appear in the next set of clips without a visual signal, and incorrectly assigns the blanks as different character IDs, P3 and P4, after the second blank. The text only model also fails to link the character in third and last clip as the same ID due to the limited information available in textual descriptions alone. This shows that the task is hard for the text-only model, while \emph{our final model learns to incorporate visual information successfully}.

\subsection{\label{sec:exp:task3} \TASKGenerate{}}

\newcommand{\midruleStatsNeww}{\cmidrule(rr){1-1} \cmidrule(rr){2-4}}

\begin {table}[t]
\begin{center}
\footnotesize
\scalebox{0.9}{
\begin{tabular}{@{}l@{\ }c@{\ \ }c@{\ \ }c@{}}
\toprule
\textbf{} & \multicolumn{3}{c}{\textbf{Per set, MAX score}}\\
\textbf{Method} & \textbf{METEOR} & \textbf{BLEU@4} & \textbf{CIDEr-D} \\
\midruleStatsNeww
Same ID & 9.22 & 1.59 & 6.31 \\
All different IDs & 8.84 & 1.38 & 6.06 \\
Ours Text-Only & 10.29 & 1.74 & 6.88 \\
Ours & 10.38 & 1.75 & 6.95 \\
\bottomrule
\end{tabular}
}
\end{center}
\caption{\textbf{\TASKGenerate{}} scores for our method on the LSMDC Test set. See Section~\ref{sec:exp:task3} for details.}
\label{tab:task3_val}
\end {table}

Finally, we evaluate our two-stage approach for the \textbf{\TASKGenerate{}} task.
One issue with directly evaluating predicted descriptions with the character IDs is that the evaluation can strongly penalize predictions that do not closely follow the ground-truth. Consider an example with ground-truth \emph{[P1] approaches [P2], and hugs [P2].} vs. prediction \emph{[P1] is approached by [P2], who hugs [P1].} As we see, direct comparison would yield a low score for this prediction, due to different phrasing which leads to different local person IDs. If we instead consider an ID permutation for the ground-truth sentence \emph{[P2] approaches [P1], and hugs [P1]}, we can get a matching bi-gram \emph{hugs [P1]} with the prediction. Thus, we consider all the possible ID permutations as references, evaluate our prediction w.r.t. all of them, and choose the reference that gives the highest BLEU@4 to compute the final scores. The caption with the best ID permutation is then used to evaluated at set level using the standard automatic scores (METEOR~\cite{meteor}, BLEU@4~\cite{bleu}, CIDEr-D~\cite{cider}).

\subsubsection{Results}

In Table~\ref{tab:task3_val}, we compare results from our captioning model  (Section~\ref{ssec:generate}) with different \TASKFillIn{} approaches to fill in the IDs on the LSMDC Test set. Our model outperforms the baseline approaches, including Ours  Text-Only model. This confirms that our \TASKFillIn{} model successfully uses visual signal to perform well, on both ground truth and predicted sentences.

\begin{figure}[t!]
\scriptsize
\begin{center}
\includegraphics[width=0.7\linewidth]{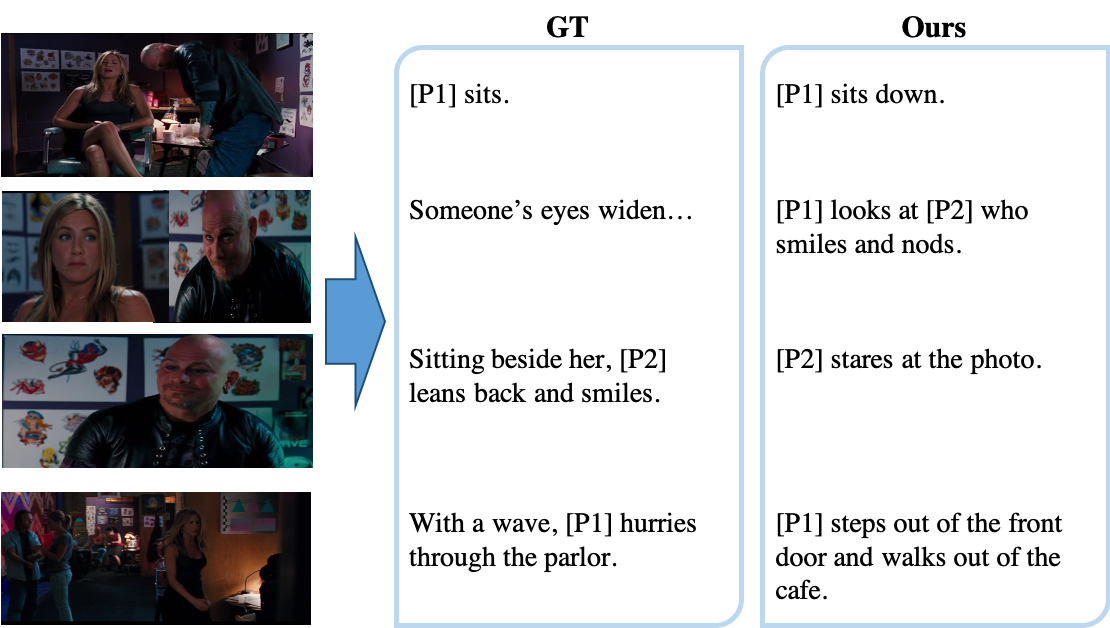}
\caption{Qualitative example for the \textbf{\TASKGenerate{}} task. P1, P2, ... are person IDs. See Section~\ref{sec:exp:task3} for details.}
\label{fig:qual_task3}
\end{center}
\vspace{-0.5cm}
\end{figure}

Figure~\ref{fig:qual_task3} provides an example output of our two-stage approach. We show the ground truth descriptions with person IDs on the left, and our generated descriptions with the predicted IDs on the right.
Our \TASKFillIn{} model consistently links [P1] as the woman who ``sits down'', ``looks at [P2]'', and ``steps out'', while [P2] as the man who ``smiles'' and ``stares'' across the clips. %

While we experiment with a fairly straightforward video description model, our two-stage approach can enable character IDs if applied to any model.

\section{\label{sec:conclusion} Conclusion}

In this work we address the limitation of existing literature on automatic video and movie description, which typically ignores the aspect of person identities. 

Our main effort in this paper is on the \TASKFillIn{} task, namely filling in the local person IDs in the given descriptions of clip sequences. We propose a new approach based on the Transformer architecture, which first learns to attend to the faces that best match the blanks, and next jointly establishes links between them (infers their IDs). Our approach successfully leverages gender information, inferred both from textual and visual cues. Human performance on the \TASKFillIn{} task shows the importance of visual information, as humans perform much better when they see the video. While we demonstrate that our approach benefits from visual features (higher ID accuracy and gender accuracy, better ID distribution), future work should focus on better ways of incorporating visual information in this task. Finally, we compare to two state-of-the-art multi-modal methods, showing a significant improvement over them.

We also show that our \TASKFillIn{} model enables us to use a two-stage pipeline to tackle the \TASKGenerate{} task. While this is a simple approach, it is promising to see that our model can handle automatically generated descriptions, and not only ground-truth descriptions. We hope that our new proposed tasks will lead to more research on bringing together person re-identification and video description and will encourage future works to design solutions that go beyond a two-stage pipeline. 

\myparagraph{Acknowledgements.}
The work of Trevor Darrell and Anna Rohrbach was in part supported by the DARPA XAI program, the Berkeley Artificial Intelligence Research (BAIR) Lab, and the Berkeley DeepDrive (BDD) Lab.

\clearpage

\section*{Supplemental Material}
\appendix

Here, we provide some technical details as well as qualitative examples and analysis. Section~\ref{sec:supp:details} provides details regarding our local person ID re-labeling, data augmentation, and accuracy metric for the \textbf{\TASKFillIn{}} task. Section~\ref{sec:supp:task2_qual} includes a qualitative comparison of our approach to \textbf{\TASKFillIn{}} and two concurrent methods, Yu et al. and Brown et al. We also discuss some failure cases. In Section~\ref{sec:supp:task2_analysis} we provide %
further insights into human performance on the \textbf{\TASKFillIn{}} task. Finally, in Section~\ref{sec:supp:task3_qual} we include \textbf{\TASKGenerate{}} scores on the validation set and more qualitative results for our approach to \TASKGenerate{} task.

\section{Technical Details}\label{sec:supp:details}

\begin{figure}[t]
\scriptsize
\begin{center}
\includegraphics[width=\linewidth]{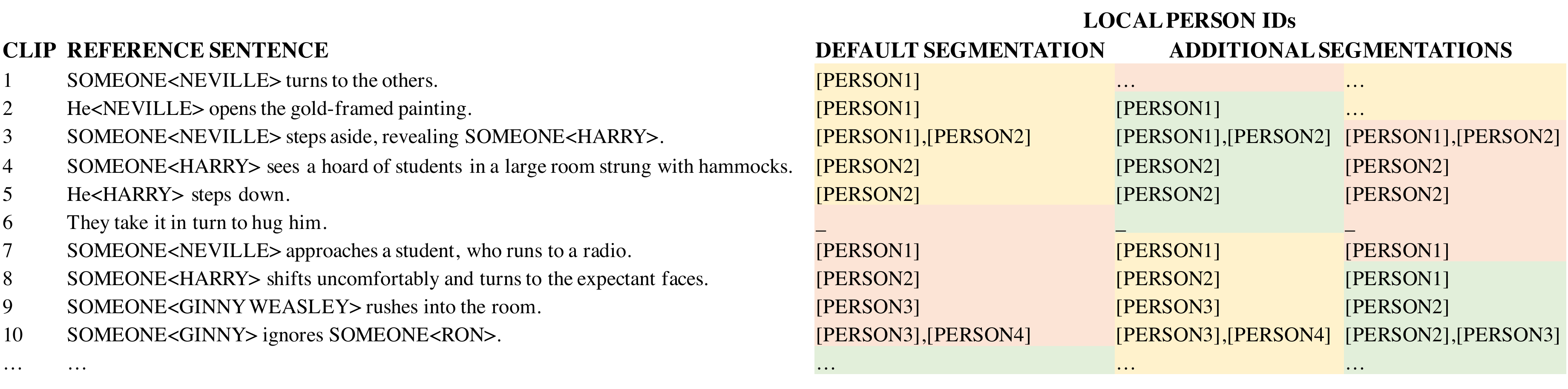}
\caption{Illustration of local person ID re-labeling and training data augmentation for the \textbf{\TASKFillIn{}} task.}
\label{fig:supp:augmentation}
\end{center}
\end{figure}

First, we illustrate our local person ID re-labeling and training data augmentation in Figure~\ref{fig:supp:augmentation}. One can see the default segmentation (consecutive sets of 5 clips) and additional segmentations, which serve for data augmentation. Each individual set of 5 clips with associated local IDs serves as a unique data point for our model.

\begin{figure}[t]
\scriptsize
\begin{center}
\includegraphics[width=\linewidth]{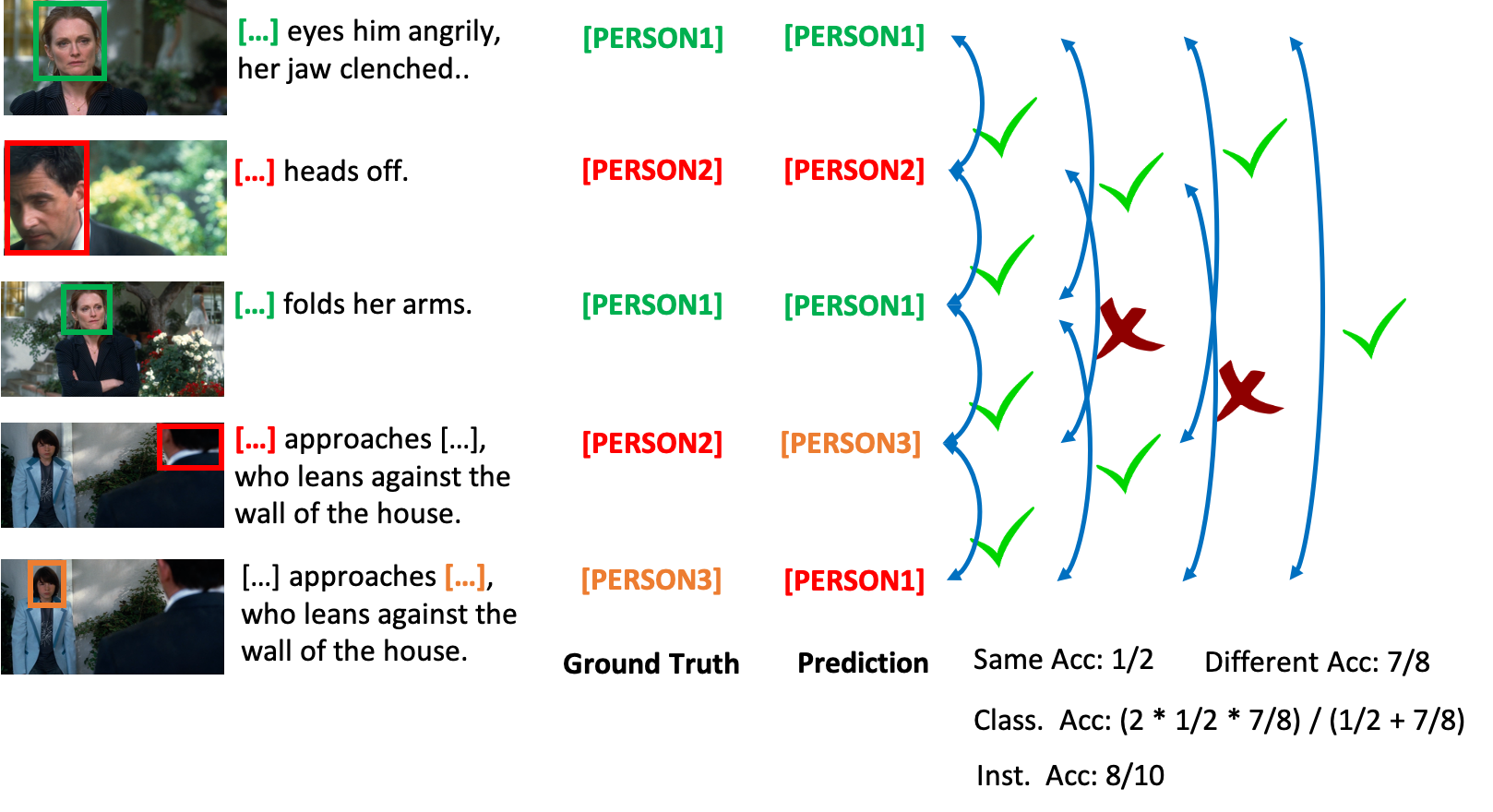}
\vspace{-0.3cm}
\caption{Illustration of the accuracy metric for the \textbf{\TASKFillIn{}} task. For each pair of blanks, we assign ``correct'' if the IDs are the same or different in \emph{both} ground truth and predictions. See Section~\ref{sec:supp:details} for more details.}
\vspace{-0.7cm}
\label{fig:supp:evaluation}
\end{center}
\end{figure}

Next, we illustrate our proposed accuracy metric (introduced in Section 5.2 of the main paper) in Figure~\ref{fig:supp:evaluation}. We consider pairwise comparisons between predicted IDs (whether they are the same or not), and compare that to ground-truth pairs. Accuracies are then computed respectively as the number of correct pairs divided by the total number of ground truth pairs with the same IDs (Same Acc), with the different IDs (Different Acc), and with all the pairs (Total Acc). In the figure, there are 2 ground truth pairwise comparisons with same IDs, 8 with different IDs, and 10 total. Note that the Same Acc and Different Acc are used to calculate the F1 score in the main paper.

\section{\textbf{\TASKFillIn{}}: Qualitative Results}\label{sec:supp:task2_qual}
\begin{figure*}[t]
\scriptsize
\begin{center}
\vspace{-5mm}
\subfloat[]{{\includegraphics[width=0.85\linewidth]{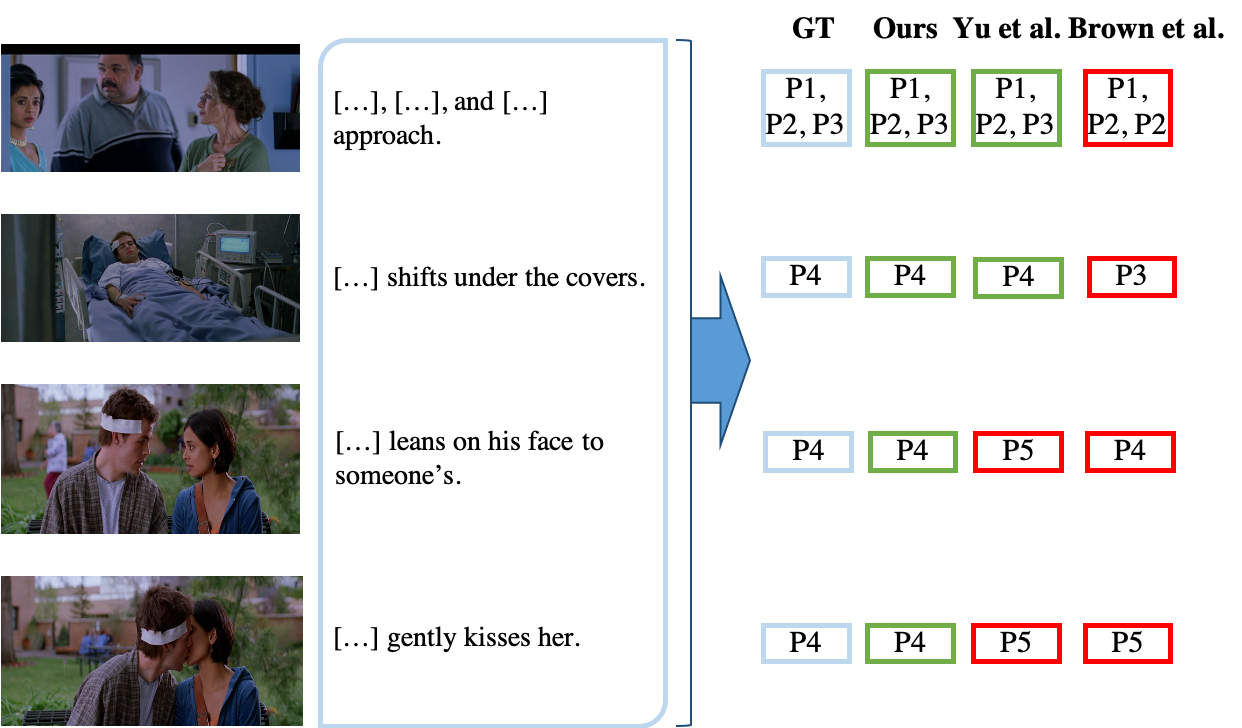} }}
\\%
\subfloat[]{{\includegraphics[width=0.85\linewidth]{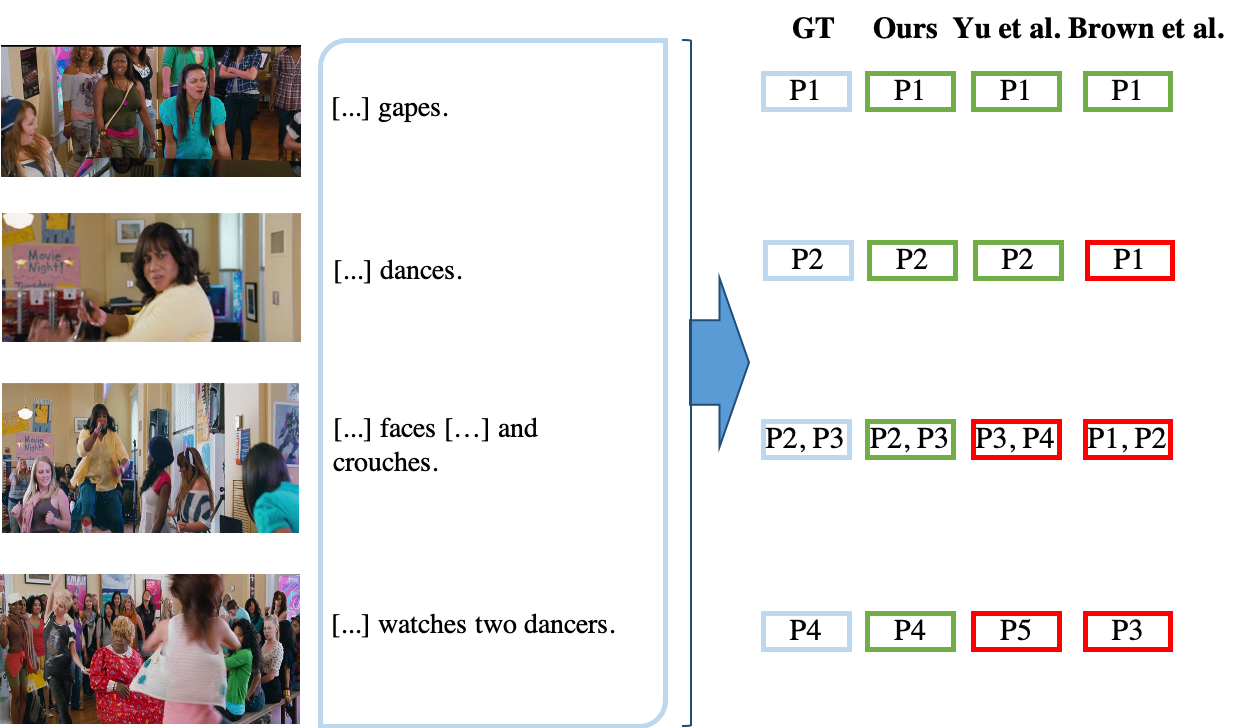} }} \\
\subfloat[]{{\includegraphics[width=0.85\linewidth]{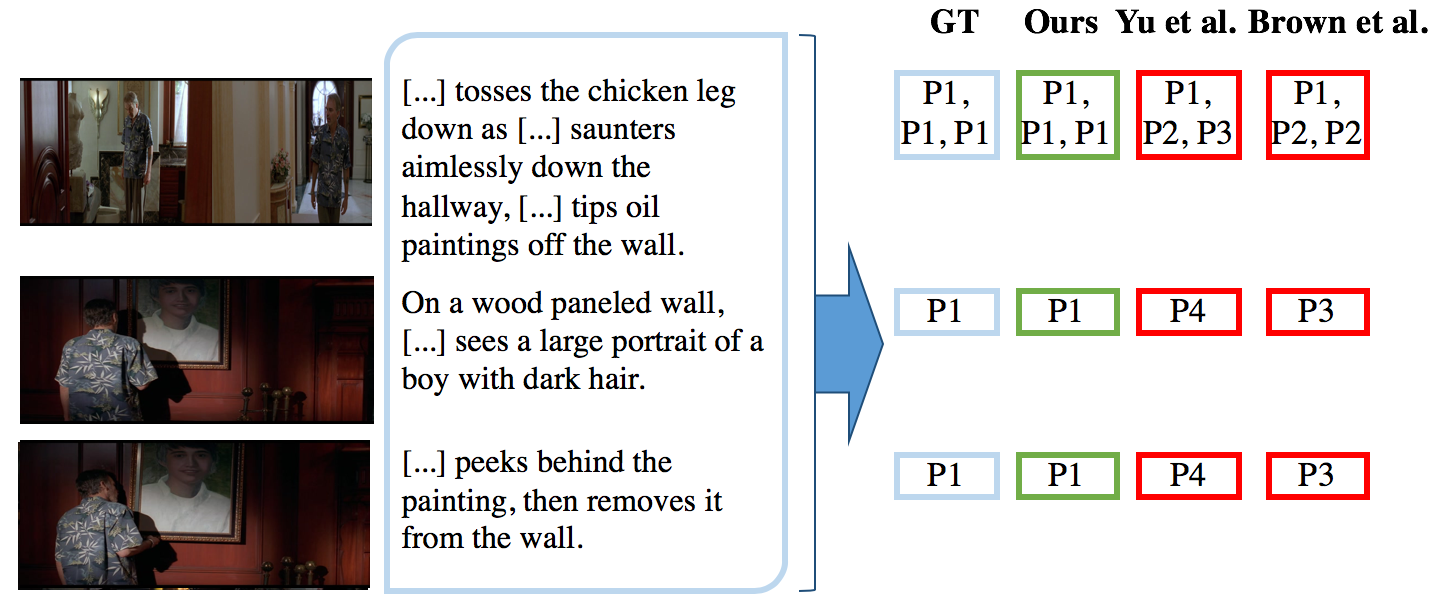} }} \\
\caption{Qualitative examples for the \textbf{\TASKFillIn{}} task, comparison between our approach and two concurrent methods. Correct/incorrect predictions are labeled with green/red, respectively.}
\label{fig:supp:task2_qual1}
\end{center}
\end{figure*}

In the qualitative examples, our model is able to recognize different characters and link the same characters across the video\footnote{Note, that we skip the clips/sentences with no blanks, therefore, sometimes resulting in less than 5 clips per set.}. In Figure~\ref{fig:supp:task2_qual1} (a), our model consistently links the character that appears from the second clip with the same ID. Likewise, we get the correct identities for video clip that involves more than two characters in Figure~\ref{fig:supp:task2_qual1} (b). On the other hand, other state of the art models either tend to predict characters that are not present in the video e.g. predicting [PERSON5] for the last sentence in Figure~\ref{fig:supp:task2_qual1} (a), or fail to correctly link to previously seen characters e.g. mixing up [PERSON4] in Figure~\ref{fig:supp:task2_qual1} (a) and [PERSON2] in Figure~\ref{fig:supp:task2_qual1} (b).
We also study if the model is possibly biased towards number of blanks in each clip. In particular, it is likely that sentence with more than one blank slot may also involve visual content with more than one character. In Figure~\ref{fig:supp:task2_qual1} (c), we show a sequence of video clips involving only one character. While our prediction identifies all the blanks as the same character, the other models struggle to do so and predict different identities. This pattern is not surprising, as we've seen that these models tend to over-predict diverse IDs (2,3,...) based on Figure~4 of the main paper.   

\begin{figure*}[t]
\scriptsize
\begin{center}
\subfloat[]{{\includegraphics[width=0.8\linewidth]{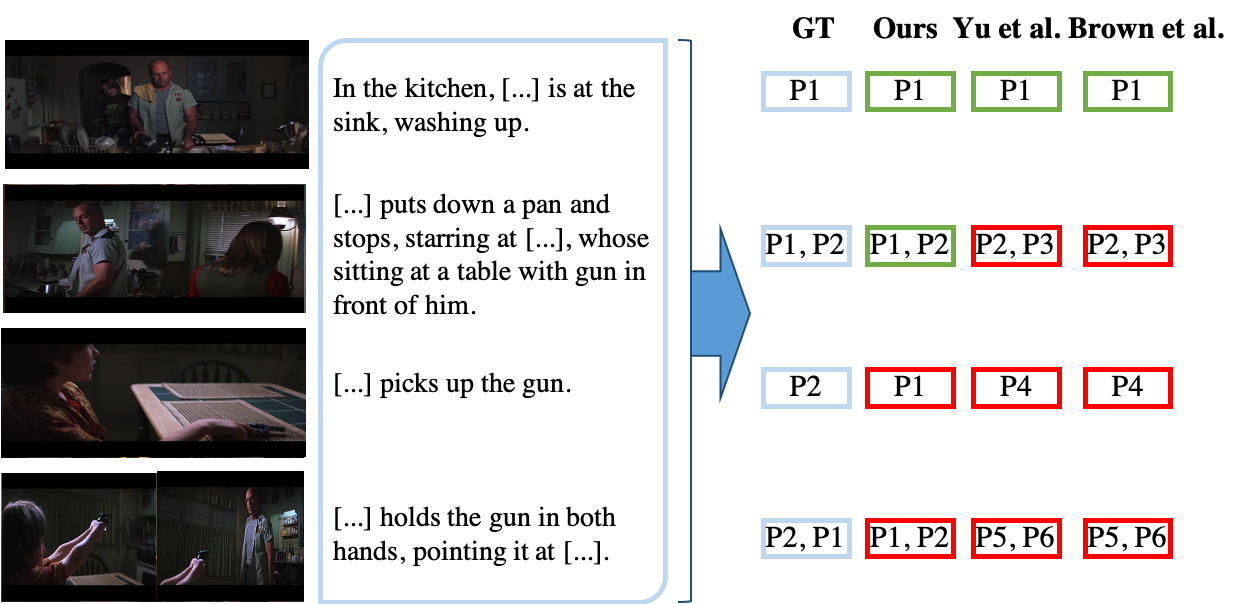} }} \\%
\subfloat[]{{\includegraphics[width=0.8\linewidth]{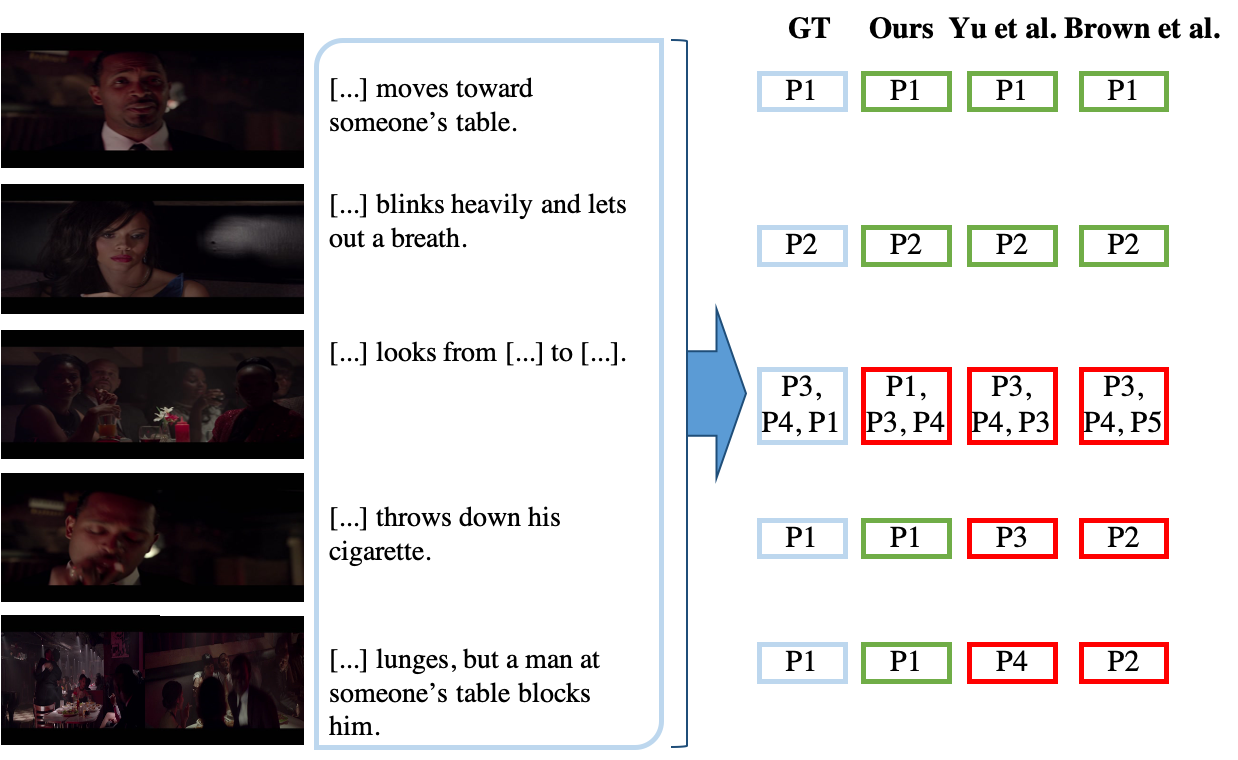} }}
\caption{Failure examples for our model on the \textbf{\TASKFillIn{}} task, comparison between our approach and two concurrent methods. Correct/incorrect predictions are labeled with green/red, respectively.}
\label{fig:supp:task2_qual2}
\end{center}
\end{figure*}

We show some failure cases in Figure~\ref{fig:supp:task2_qual2}. In the second clip and the last clip of Figure~\ref{fig:supp:task2_qual2} (a), we fail to identify which character is holding the gun and who they are pointing at. This results in the swapped character IDs. Note, that we still limit our predictions to two characters, while other methods predict more characters. Our model also struggles to correctly link previously seen characters in clips with a crowd of people. In Figure~\ref{fig:supp:task2_qual2} (b), the model incorrectly links the characters within in the third clip due to a large crowd; however, it still links the last two IDs as the same person in the first clip, which matches the ground truth labels.

\section{{\TASKFillIn{}}: Additional Analysis}\label{sec:supp:task2_analysis}

In Section 5.3.2 of the main paper, we have presented human performance on the \textbf{\TASKFillIn{}} task, measured as a median accuracy across three annotators. It is also worth looking at the maximum accuracy across three workers. We include that in Table~\ref{tab:supp:task2_test}. As we see, the numbers are substantially higher, if we consider the upper-bound accuracy across the workers. When seeing the video, humans can get up to {96\%} accuracy. We analyze the cases when none of the three annotators were able to get the correct person IDs, and find that most of the time that happens (a) in more complex scenes (with multiple participants), (b) in symmetrical cases like ``[...] and [...] walk in'', (c) in  ambiguous cases, such as ``[...] gives [...] a look'', where it may be hard to tell which of the two persons was meant, etc.

\begin{table}[t]
\begin{center}
\small
\begin{tabular}{@{}l@{\ }c@{}}
\toprule
\textbf{Method} & \textbf{Inst Acc} \\
\midrule
Human w/o video (median) & 70.0 \\
Human (median) & 87.0 \\
\midrule
Human w/o video (max) & 85.1 \\
Human (max) & 96.0 \\
\bottomrule
\end{tabular}
\end{center}
\caption{\textbf{\TASKFillIn{}}: median and maximum human performance over 200 random sets of clips (Test set).}
\vspace{-0.5cm}
\label{tab:supp:task2_test}
\end {table}

\section{{\TASKGenerate{}}: Additional Results}\label{sec:supp:task3_qual}

We show results from our captioning model and other baselines on the validation set in Table~\ref{tab:supp:task3_val}. We use the same metric in Table 5 of the main paper. 

\begin {table}[h]
\begin{center}
\small
\begin{tabular}{@{}l@{\ }c@{\ \ }c@{\ \ }c@{}}
\toprule
\textbf{} & \multicolumn{3}{c}{\textbf{Per set, MAX score}}\\
\textbf{Method} & \textbf{METEOR} & \textbf{BLEU@4} & \textbf{CIDEr-D} \\
\midrule
Same ID & 9.41 & 1.57 & 7.03 \\
All different IDs & 9.11 &  1.36 & 7.00 \\
Ours Text-Only & 10.53 & 1.77 & 7.73 \\
Ours & 10.68 & 1.80 & 7.77  \\
\bottomrule
\end{tabular}
\end{center}
\caption{\textbf{\TASKGenerate{}} scores for our method on the LSMDC validation set.}
\vspace{-0.5cm}
\label{tab:supp:task3_val}
\end {table}

\begin{figure*}[t]
\scriptsize
\begin{center}
\subfloat[]{{\includegraphics[width=0.9\linewidth]{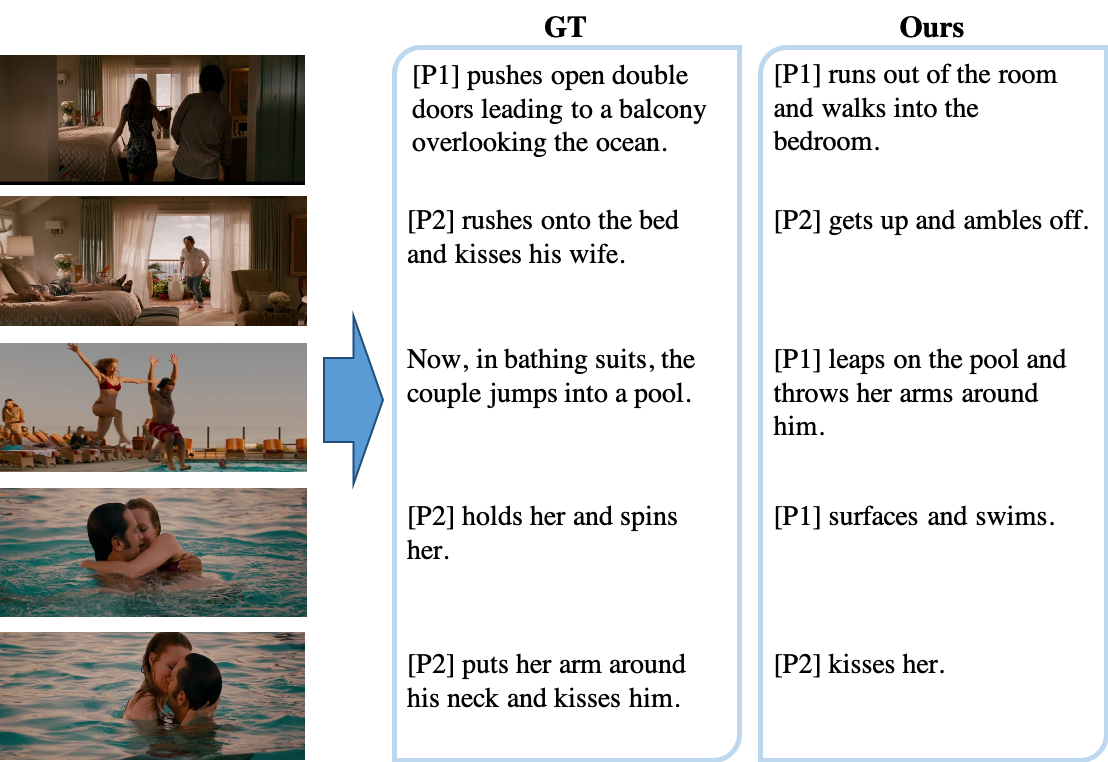} }} \\%
\subfloat[]{{\includegraphics[width=0.9\linewidth]{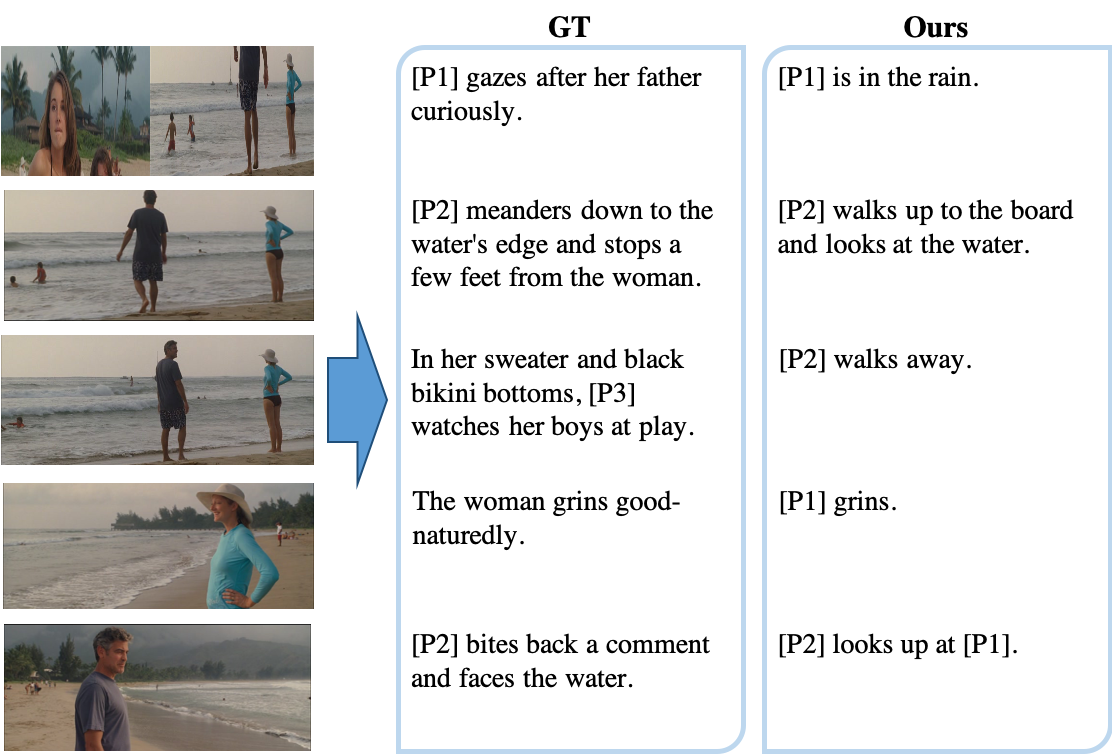} }} \\

\caption{Qualitative examples of our approach on the \textbf{\TASKGenerate{}} task.}
\label{fig:supp:task3_qual}
\end{center}
\end{figure*}

Finally, in Figure~\ref{fig:supp:task3_qual}, we show descriptions generated by our baseline model with predicted character IDs. Overall, our \TASKFillIn{} model correctly links relevant activity to the character IDs. In Figure~\ref{fig:supp:task3_qual} (a), PERSON1 is identified as the woman who ``walks into the bedroom" and ``throws her arms around him", and PERSON2 is the man who ``kisses her" in the pool. Likewise, PERSON1 and PERSON2 are consistently linked as the woman and man using our approach in Figure~\ref{fig:supp:task3_qual} (b). However, we do acknowledge that our two-stage pipeline approach is not perfect; our model does not identify all the characters in the video, e.g. there is no ID for the girl who ``gazes after her father" in the first clip in Figure~\ref{fig:supp:task3_qual} (b). We leave improving the description quality with character identification to future work.

\clearpage

%
%
\bibliographystyle{splncs04} 
\bibliography{biblioLong,egbib}
\end{document}